\documentclass[11pt]{article}

\usepackage[final]{acl}

\usepackage{times}
\usepackage{latexsym}

\usepackage[T1]{fontenc}

\usepackage[utf8]{inputenc}

\usepackage{microtype}

\usepackage{inconsolata}
\usepackage{booktabs}
\usepackage{multirow}
\usepackage{array}
\usepackage{xcolor}
\usepackage{url}
\usepackage{fvextra}

\usepackage{graphicx}

\newcommand{\benchmark}{Food4All}
\newcommand{\deltafmt}[2]{\textcolor{#1}{\footnotesize #2}}
\newcommand{\traitcell}[3]{#1\,\deltafmt{#2}{#3}}
\DefineVerbatimEnvironment{PromptBlock}{Verbatim}{fontsize=\scriptsize,breaklines=true,breakanywhere=true}

\title{Food4All: An Agentic Framework and Benchmark for Food Resource Navigation with Adaptive User Understanding}

\author{
 \textbf{Yiyang Li\textsuperscript{1}},
 \textbf{Weixiang Sun\textsuperscript{1}},
 \textbf{Tianyi Ma\textsuperscript{1}},\\
 \textbf{Kaiwen Shi\textsuperscript{1}},
 \textbf{Zheyuan Zhang\textsuperscript{1}},
 \textbf{Yanfang Ye\textsuperscript{1\textdagger}}
\\
 \textsuperscript{1}University of Notre Dame \quad \textsuperscript{\textdagger}Corresponding Author \\
 \texttt{\{yli62, yye7\}@nd.edu} 
}

\begin{document}
\maketitle

\begin{abstract}
Food assistance referral requires conversational agents to translate underspecified, often noisy help-seeking dialogues into locally valid resource recommendations. We present \textbf{\benchmark}, an agentic food-resource referral framework and benchmark grounded in 686 structured Indiana food resources. \benchmark{} couples a food-specific search tool with 300 multi-turn evaluation tasks spanning single food needs, composite cases with access or document constraints, and five non-ideal user interaction traits: unreasonable demands, rambling responses, impatience, incomplete answers, and inconsistent information. We evaluate six Large Language Models (LLMs) on requirement grounding, resource retrieval, final referral correctness, and interaction efficiency. Although the strongest model achieves 96.33\% referral accuracy, our diagnostics reveal persistent failures in grounding schedule, eligibility, intake, and document constraints, as well as failures to preserve valid retrieved resources in the final recommendation. Trait-level analysis further shows that different non-ideal behaviors stress different parts of the referral pipeline. \benchmark{} provides a controlled testbed for studying tool-calling agents in constraint-sensitive food assistance referral under realistic user interaction challenges. Our codebase is available \href{https://github.com/Yiyang-Ian-Li/Food4All}{here}.
\end{abstract}

\section{Introduction}

In the United States, access to food assistance is a large-scale and persistent public need. Food insecurity affects millions of households and is closely tied to other social determinants of well-being, such as poverty, housing instability, and health. In 2024, 47.9 million people lived in food-insecure households, meaning that they had limited or uncertain access to adequate food at some point during the year~\cite{usda2026foodsecurity}. In response to these needs, public nutrition programs, such as the Supplemental Nutrition Assistance Program (SNAP), served an average of 41.7 million participants per month in fiscal year 2024~\cite{usda2025snap}. At the same time, food assistance is delivered through a complex ecosystem of public programs and community-based providers, including SNAP, WIC\footnote{\url{https://www.fns.usda.gov/wic}}, food banks, food pantries, and community meal sites. For example, Feeding America describes its nationwide network as including more than 250 food banks and 60,000 agency partners, food pantries, and meal programs~\cite{feedingamerica_network}. The scale of need, the diversity of available programs, and the locality-specific nature of food assistance make referral and navigation a critical part of food access infrastructure.

However, significant challenges remain in making this infrastructure accessible to those who need it. First, providing high-quality, around-the-clock food assistance consultation support is labor-intensive and costly~\cite{wa2112023expansion}. Second, help-seeking populations often include a large proportion of socially vulnerable and marginalized individuals~\cite{kreuter2012use, thompson2016promoting}, many of whom may have limited digital literacy~\cite{anuyah2023exploring, mcclain_bishop_2026_digital_divides_us}. Research suggests that these users face barriers when using web-based information retrieval tools and often prefer conversational interaction when seeking information and support~\cite{bickmore2016improving, kocielnik2020harborbot, kocielnik2021can}.

Recent advances in Large Language Models (LLMs) \cite{ye2025llms4all} based tool-calling agents have created new opportunities for conversational food resource navigation~\cite{schick2023toolformer, alkhouli2025confetti}. In a typical tool-calling agent workflow, the agent interacts with a user to collect relevant information, invokes tools to query external databases, and then utilizes the retrieved information~\cite{qin2024toolllm, patil2024gorilla}. However, real-world food assistance referral poses substantial challenges for such agents. First, before calling a resource-search tool, an agent must accurately understand the user's needs and constraints, including the type of food assistance sought, geographic constraints, schedule, program-qualification facts, intake method, and document availability. Second, real users often do not interact with agents in an idealized or fully cooperative manner~\cite{shim2025non}. Some users may make unreasonable demands that arise from a legitimate food need, give off-topic responses, answer incompletely, become impatient, or present self-contradictory information. These behaviors make requirement elicitation and food resource matching considerably more difficult.

Existing benchmarks for tool-calling agents primarily evaluate abilities such as API selection, multi-step tool use, and task completion in structured environments~\cite{li2023api, qin2024toolllm, yao2024tau, lu2025toolsandbox}. However, they do not adequately capture the interactional complexity and domain-specific constraints of real-world social service referral. To systematically study agent capabilities and behavioral patterns in this setting, we introduce \textbf{\benchmark}, a food assistance referral framework with a benchmark layer for evaluating conversational tool use. \benchmark{} is built from 686 filtered resources covering four food service types: pantry or groceries, prepared meals, home-delivered meals, and baby food or formula. The benchmark includes 300 tasks from 50 seed cases, with 25 single cases and 25 composite cases that add firm intake or document constraints to one food need. It also includes five non-ideal interaction traits: \textit{(i) unreasonable demands}, \textit{(ii) rambling responses}, \textit{(iii) impatience}, \textit{(iv) incomplete answers}, and \textit{(v) inconsistent information}. Based on this benchmark, we investigate the following research questions:

\begin{itemize}
    \item \textbf{RQ1:} Can agents elicit and ground users' requirements into successful resource retrieval and referral?
    \item \textbf{RQ2:} Where do failures arise, including parameter-grounding errors and resources that are retrieved but not recommended?
    \item \textbf{RQ3:} How do non-ideal user interactions affect referral performance and interaction efficiency?
\end{itemize}

Our experiments show that current LLM agents can use the \benchmark{} framework, but the benchmark layer separates several abilities that are often conflated in tool-use evaluation. A model may elicit the right facts but over-filter the search call, retrieve an acceptable resource but fail to copy it into the final referral, or handle ordinary users while breaking down when a user gives a same-turn contradiction. Incomplete answers mainly add elicitation cost, whereas inconsistent answers require the agent to pause and repair the information state before searching. Unreasonable demands test a different behavior: the agent must recover the underlying food need without turning the impossible request into a hard search constraint.

Our contributions are twofold. First, we build a food-resource referral framework that normalizes real Indiana 211 food resources into a structured search interface usable by LLM agents. Second, we construct an interaction-centered benchmark that evaluates whether agents can elicit, ground, retrieve, and recommend food resources under non-ideal user interactions.

\section{Related Work}

\subsection{Tool-Calling Agent Evaluation}

Recent work increasingly evaluates LLMs as agents that can interact with external tools \cite{chen2025obvious,chen2025clear}. Existing benchmarks study whether agents can select APIs, construct valid function calls, compose multiple tool invocations, and complete tasks in executable environments~\cite{li2023api, qin2024toolllm, patil2024gorilla}. More recent benchmarks extend this evaluation to interactive settings, where agents must converse with simulated users, follow domain policies, maintain state, and decide when to call tools~\cite{alkhouli2025confetti, yao2024tau, lu2025toolsandbox}. While these benchmarks have advanced the evaluation of tool-calling agents, they are not designed for social service referral, where agents must elicit missing requirements, ground partially specified needs into structured search parameters, and match users to resources under specific requirements and constraints.

\begin{figure*}[!ht]
    \centering
    \includegraphics[width=\linewidth]{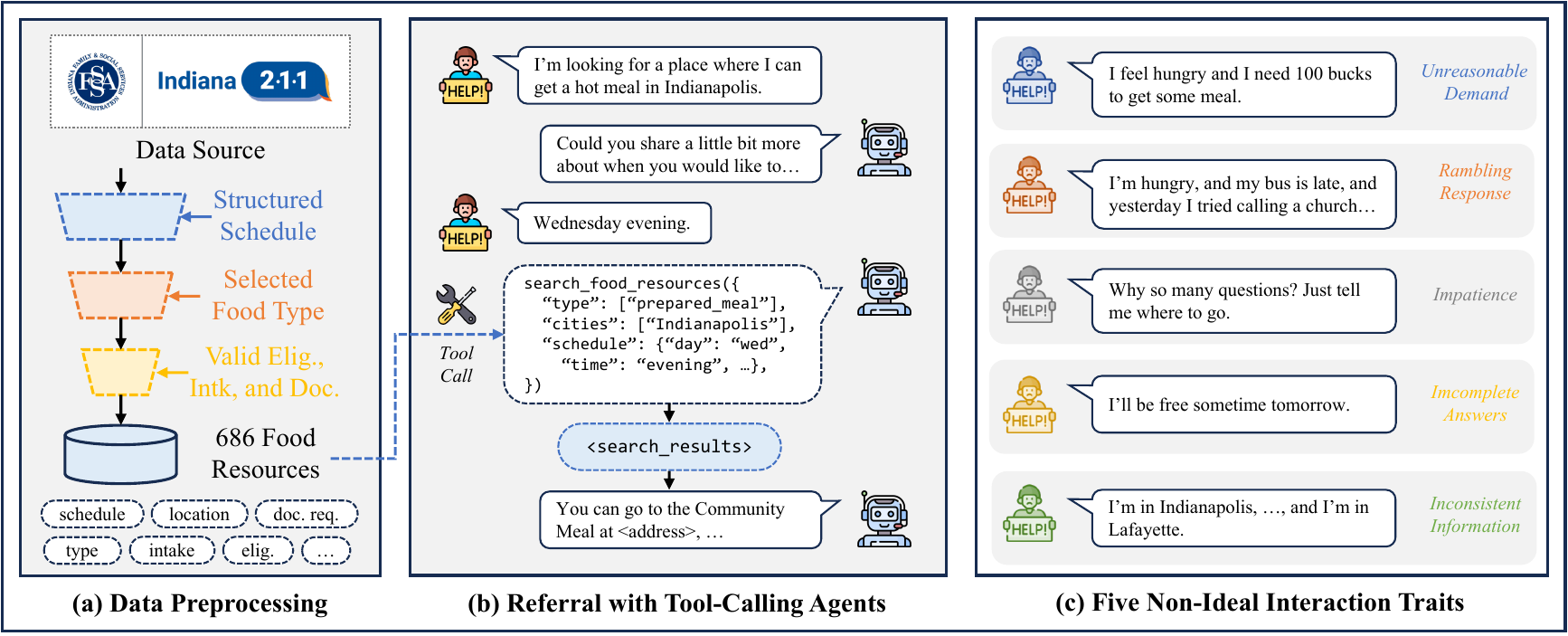}
    \caption{Overview of \benchmark. (a) We construct a food-resource database from Indiana social service data by extracting structured schedules, selecting food-related resource types, and filtering resources with valid eligibility, intake, and documentation information. (b) In each task, a tool-calling agent interacts with a simulated help-seeker, elicits food-related needs and constraints, calls the food-resource search tool, and recommends retrieved resources. (c) The benchmark evaluates agents under normal users and users with five non-ideal interaction traits: unreasonable demands, rambling responses, impatience, incomplete answers, and inconsistent information.}
    \label{fig:method}
    \vspace{-10pt}
\end{figure*}

\subsection{Conversational Social Service Navigation}

Social service access often depends on information-and-referral processes that connect people to resources matching their needs and circumstances. Prior work suggests that web-based resource search can be difficult for users who face digital literacy, language, or socioeconomic barriers, while conversational interfaces can make information seeking more understandable, personal, and supportive~\cite{bickmore2016improving, kocielnik2020harborbot, kocielnik2021can}. Various AI-driven approaches \cite{zhao2023self,chen2018droideye,ye2009intelligent,ju2022grape,qian2022co,zhao2021multi,ju2023graphpatcher} can be used for soccial service naviation, while recent work explores chatbots for social-needs screening and resource sharing, demonstrating the promise of conversational approaches in healthcare and community settings~\cite{sezgin2024chatbot}. However, this work primarily studies system design, user experience, and feasibility rather than systematic benchmark evaluation. \benchmark{} instead evaluates LLM agents for realistic, interaction-sensitive food resource referral.

\section{Food4All System and Benchmark}

\subsection{Overview}

\benchmark{} is both a food-resource referral framework and an evaluation benchmark for conversational tool-calling agents.
At inference time, an agent must interact with a help-seeking user, elicit missing food, location, schedule, eligibility, intake, and document facts, translate those facts into structured search parameters, retrieve candidate resources, and produce a final referral.
To support both agent-facing search and controlled evaluation, \benchmark{} combines four components: a food-specific resource index, a structured food-search tool, deterministic case specifications with acceptable resources, and a runtime protocol for multi-turn agent-user interaction.

We construct the benchmark from an Indiana 211 source snapshot and retain local Indiana resources whose food-related fields can be normalized into a structured search index.
The processed food index is further filtered into an evaluation-ready subset with parseable schedule windows, known food types, known eligibility tags, parseable intake methods, parseable document requirements, city and ZIP metadata, and at least one contact channel.
We expose this index through a single \texttt{search\_food\_resources} tool that applies hard filters over food type, location, eligibility, schedule, intake method, and available documents.
Finally, we construct simulated users from thin user cards whose openings mention only the food need and geographic requirement while hidden facts define the remaining constraints. This design makes the benchmark automatically scorable while preserving core referral challenges.
Users describe food needs in ordinary language, acceptable resources are not always unique, and the agent must avoid adding unsupported filters when a user has no special access or document requirement.

\subsection{Resource Collection and Filtering}

We collect resource records from an Indiana 211 source snapshot.
The source contains deduplicated service-site records with agency and site names, service names, taxonomy labels, service areas, addresses, contact channels, schedule text, access instructions, eligibility text, and required documents.
We first retain local Indiana food resources and normalize food-specific fields into a processed index with 2,422 resources across 92 county or service-area labels.

Because \benchmark{} evaluates structured tool use rather than open-ended web search, we build a filtered index whose tool arguments correspond to reliable fields.
We retain resources whose schedules can be parsed into structured windows, whose food type is one of the benchmark food types, whose eligibility tags are known, whose intake method is parseable, whose document requirements are known rather than ambiguous, and whose metadata is sufficient for search and referral.

The final benchmark index contains 686 food resources across 86 county labels, 234 cities, and 307 ZIP codes.
It covers four benchmark food types: pantry or groceries, prepared meals, home-delivered meals, and baby food or formula.
Eligibility facts are represented as coarse tags such as service-area residency, low income, senior status, child or student, pregnancy or WIC, homebound or disabled, veteran status, and open eligibility.
Document requirements are represented as explicit tags such as photo ID, proof of address, proof of income, lease, utility bill, birth certificate, insurance card, social security document, or no required document.
Table~\ref{tab:benchmark-stats} summarizes the resulting benchmark.

\begin{table}[t!]
\centering
\small
\begin{tabular}{lr}
\toprule
\textbf{Statistic} & \textbf{Value} \\
\midrule
Evaluation-ready resources & 686 \\
Food service types & 4 \\
Cities & 234 \\
ZIP codes & 307 \\
Indiana county labels & 86 \\
Seed cases & 50 \\
Eval samples & 300 \\
Samples per trait & 50 \\
Acceptable resources per case & 1--3 \\
Single / composite seed cases & 25 / 25 \\
\bottomrule
\end{tabular}
\caption{Summary of the filtered food resource index and evaluation cases.}
\label{tab:benchmark-stats}
\vspace{-10pt}
\end{table}

\subsection{Tool and Agent Design}

\benchmark{} exposes a single resource-search tool, \texttt{search\_food\_resources}, and wraps it with a general function-calling referral agent.
The tool is designed to support real food-resource lookup while also testing whether agents can translate conversationally elicited food-referral requirements into structured search parameters.
Its arguments include optional location fields, \texttt{service\_tracks}, \texttt{food\_service\_types}, \texttt{eligibility\_tags}, \texttt{intake\_methods}, \texttt{available\_documents}, \texttt{schedule}, \texttt{include\_appointment\_only}, and \texttt{limit}.
The location fields are \texttt{counties}, \texttt{cities}, and \texttt{zipcodes}, with valid ZIP codes taking precedence over cities and counties.
The \texttt{schedule} argument supports day plus daypart, 24-hour requirements, and appointment-only resources.
The \texttt{available\_documents} argument represents documents the user can provide, and a resource matches when its concrete document requirements are a subset of those documents.

Every non-empty argument is interpreted as a hard filter.
Multiple values within the same argument are treated as alternatives.
This design makes tool calls diagnostic: if no acceptable resource is retrieved, we can inspect whether the agent over-filtered by adding unsupported constraints, selected the wrong food type, used the wrong location, or searched with the wrong schedule.

The agent receives natural-language instructions, can ask follow-up questions, and can call the food-search tool up to three times before returning a strict JSON referral.
The tool returns a list of matching resources, including \texttt{resource\_id}, \texttt{resource\_name}, \texttt{agency\_name}, service track, food service types, eligibility tags, city, ZIP code, contact fields, schedule status, intake methods, document requirements, and eligibility text.
The full tool schema shown to agents is provided in Appendix~\ref{sec:appendix-details}.

\begin{table}[t]
\centering
\small
\begin{tabular}{>{\raggedright\arraybackslash}p{0.30\columnwidth}>{\raggedright\arraybackslash}p{0.58\columnwidth}}
\toprule
\textbf{Trait} & \textbf{Definition} \\
\midrule
Normal & The user is cooperative and answers questions clearly. \\
\midrule
Unreasonable demand & The user pushes for an unrealistic outcome, such as grocery cash or instant delivery, while the underlying food need remains recoverable. \\
\midrule
Rambling responses & The user answers the question but surrounds the answer with side comments or off-topic conversation. \\
\midrule
Impatience & The user is rushed or frustrated and resists repeated clarification, while still providing enough information for the conversation to continue. \\
\midrule
Incomplete answers & The user gives an unclear or partial answer the first time a fact is requested, and provides the fact only after follow-up. \\
\midrule
Inconsistent information & The user gives a self-contradictory answer about a required fact in one reply, requiring the agent to notice and resolve the inconsistency. \\
\bottomrule
\end{tabular}
\caption{Definitions of simulated-user interaction traits.}
\label{tab:trait-definitions}
\vspace{-10pt}
\end{table}

\subsection{User and Ground-Truth Construction}

Each seed case is constructed from a candidate resource and a set of hard constraints that retrieve one to three acceptable resources.
A single case contains the core food-referral constraints: food service type, service track, eligibility facts, location requirement, and schedule requirement.
A composite case is still one food problem, but it adds a firm access, intake, or document constraint.
Composite cases therefore test whether agents can distinguish positive firm constraints from the absence of special requirements.

For each candidate, we enumerate possible location requirements from county, city, and ZIP fields and schedule requirements from structured schedule windows.
We retain a case only when applying the hidden constraints retrieves between one and three acceptable resources.
This criterion reflects referral practice more naturally than a single-answer setup while preserving automatic scoring: an agent receives credit if it retrieves or recommends any acceptable resource.

The final benchmark contains 50 seed cases and 300 evaluation samples.
There are 25 single seed cases and 25 composite seed cases.
Each seed case is expanded across six interaction traits, yielding 150 single-case samples and 150 composite-case samples.
Location requirements use ZIP codes in 18 seed cases, cities in 16 seed cases, and counties in 16 seed cases.
All seed cases use a day plus daypart schedule requirement.

After deterministic case generation, we create a thin simulated-user card for each seed case.
The opening message mentions only the food need and geographic requirement.
Hidden fields contain the location requirement, the food need summary, food type, eligibility facts, schedule requirement, and any firm intake or document constraint.
The simulated user is not asked to reveal a resource ID or database label; it answers questions in ordinary language from the hidden facts.

Each evaluation sample is assigned one interaction trait.
The traits include one normal baseline and five non-ideal interaction traits: unreasonable demands, rambling responses, impatience, incomplete answers, and inconsistent information.
The benchmark is balanced across traits, with 50 samples per trait.
Table~\ref{tab:trait-definitions} defines the traits.

\subsection{Runtime Protocol}

At runtime, each episode begins with the simulated user's opening message. The agent receives instructions describing the referral task, the \texttt{search\_food\_resources} tool schema, and the required final-answer format. The agent may ask follow-up questions and may call the tool up to three times. Each episode allows at most eight user turns.

The agent is instructed to clarify the food help type, one location constraint, schedule, access or intake constraints, document availability, and relevant program-qualification facts before searching.
It is also instructed to ask qualification questions in ordinary language rather than asking the user for eligibility tags.
An episode terminates when the agent produces valid final JSON with a \texttt{recommendations} list or when the turn limit is reached.
Final recommendations must copy fields from selected tool results.
If no suitable resource is found, the required output is an empty recommendation list.
This protocol allows us to evaluate not only whether the agent retrieves an acceptable resource, but also whether it faithfully converts retrieved resources into final user-facing referrals.
The full agent instructions and simulated-user prompts are provided in the appendix.

\begin{table*}[t]
\centering
\small
\begin{tabular*}{\textwidth}{@{\extracolsep{\fill}}lcccccccc@{}}
\toprule
\textbf{Model} & \multicolumn{2}{c}{\textbf{Overall}} & \multicolumn{3}{c}{\textbf{Single}} & \multicolumn{3}{c}{\textbf{Composite}} \\
\cmidrule(lr){2-3}\cmidrule(lr){4-6}\cmidrule(l){7-9}
& \textbf{Ref.} & \textbf{Retr.} & \textbf{Ref.} & \textbf{Retr.} & \textbf{Turns} & \textbf{Ref.} & \textbf{Retr.} & \textbf{Turns} \\
\midrule
GPT-4.1 & \underline{90.33} & 90.33 & 90.00 & 90.00 & \textbf{3.17} & \underline{90.67} & 90.67 & \underline{3.38} \\
GPT-4.1-mini & 88.33 & 88.33 & 89.33 & 89.33 & 4.62 & 87.33 & 87.33 & 5.04 \\
GPT-OSS-120B & 88.00 & 89.33 & \underline{92.00} & 93.33 & \textbf{3.17} & 84.00 & 85.33 & \textbf{3.31} \\
DeepSeek-V4-Flash & \textbf{96.33} & \textbf{100.00} & \textbf{98.67} & \textbf{100.00} & 5.71 & \textbf{94.00} & \textbf{100.00} & 5.99 \\
Qwen3-235B & 68.67 & \underline{94.33} & 73.33 & \underline{97.33} & 4.00 & 64.00 & \underline{91.33} & 4.29 \\
Llama-3.3-70B & 31.00 & 35.33 & 31.33 & 34.00 & 3.73 & 30.67 & 36.67 & 4.11 \\
\bottomrule
\end{tabular*}
\caption{Referral and retrieval hit rates by model and case type. All values except turns are percentages. Ref. and Retr. denote referral and retrieval hit. Turns is the average number of user turns. Bold and underline mark the best and second-best results in each column.}
\label{tab:main-results}
\end{table*}

\begin{table*}[t]
\centering
\small
\begin{tabular*}{\textwidth}{@{\extracolsep{\fill}}lccccccc@{}}
\toprule
\textbf{Model} & \textbf{Ret-Ref} & \textbf{Svc.} & \textbf{Loc.} & \textbf{Sch.} & \textbf{Elig.} & \textbf{Intk.} & \textbf{Doc.} \\
\midrule
GPT-4.1 & 0.00 & 17.24 & 3.45 & 37.93 & 31.03 & 84.62 & 80.00 \\
GPT-4.1-mini & 0.00 & 2.86 & 0.00 & 25.71 & 11.43 & 5.56 & 33.33 \\
GPT-OSS-120B & 1.33 & 46.88 & 65.62 & 53.12 & 78.12 & 88.89 & 81.82 \\
DeepSeek-V4-Flash & 3.67 & -- & -- & -- & -- & -- & -- \\
Qwen3-235B & 25.67 & 23.53 & 5.88 & 64.71 & 82.35 & 92.31 & 80.00 \\
Llama-3.3-70B & 4.33 & 96.91 & 99.48 & 99.48 & 99.48 & 97.62 & 100.00 \\
\bottomrule
\end{tabular*}
\caption{Diagnostic errors. Ret-Ref is retrieval hit minus referral hit in percentage points. Svc., Loc., Sch., and Elig. are the percentages of retrieval-miss cases where no executed tool call matched the hidden food type, location, schedule, or eligibility field. Intk. and Doc. are computed only over retrieval-miss cases whose hidden constraints include intake or document requirements. Dashes indicate that the model had no retrieval misses.}
\label{tab:error-results}
\end{table*}

\section{Experiments}

\subsection{Experimental Setup and Metrics}

We evaluate tool-calling agents on six backbone LLMs: GPT-4.1, GPT-4.1-mini~\cite{openai2025gpt41}, GPT-OSS-120B~\cite{openai2025gptoss}, DeepSeek-V4-Flash, Qwen3-235B~\cite{yang2025qwen3}, and Llama-3.3-70B~\cite{meta2024llama33}. All runs use \texttt{gpt-4.1-mini} as the simulated user model, with a maximum of eight user turns and at most three calls to \texttt{search\_food\_resources} per episode.

We report three groups of metrics. First, performance metrics measure whether agents retrieve and recommend acceptable resources. \textbf{Retrieval hit} indicates whether any acceptable resource appears in any tool result returned during the episode, serving as an outcome-level measure of requirement grounding and tool use. \textbf{Referral hit} indicates whether the final recommendation contains an acceptable resource ID, measuring whether the agent converts retrieved information into a correct referral. Second, diagnostic error metrics decompose failures into \textbf{Ret-Ref}, the gap between retrieval and referral hit, and field-level retrieval errors for food type, location, schedule, eligibility, intake, and document constraints. Third, interaction metrics measure efficiency using the number of user turns and executed tool calls per episode.

\subsection{RQ1: Requirement Grounding and Referral Performance}

Table~\ref{tab:main-results} reports retrieval and referral hit rates across all cases, single cases, and composite cases. DeepSeek-V4-Flash retrieves an acceptable resource in every case and achieves the highest referral hit at 96.33\%, while GPT-4.1 reaches 90.33\% referral hit with fewer turns. GPT-4.1-mini and GPT-OSS-120B are close behind, showing that the tool interface is usable beyond the largest proprietary model when the relevant constraints are elicited.

Composite cases are not multi-need cases in \benchmark{}; they are single food problems with additional firm intake or document constraints. They are nevertheless harder for several models: GPT-OSS-120B drops from 92.00\% referral hit on single cases to 84.00\% on composite cases, and Qwen3-235B drops from 73.33\% to 64.00\%. The drop isolates a practical failure mode: models can identify the food need and location but lose access or document constraints when forming the search query or final referral. The turn columns show a separate tradeoff. DeepSeek-V4-Flash is the most accurate but uses almost six turns on average, while GPT-4.1 and GPT-OSS-120B are more concise but miss more constraints.

\subsection{RQ2: Diagnostic Error Analysis}

To understand where referral failures arise, Table~\ref{tab:error-results} decomposes errors into retrieval-stage and post-retrieval failures. Ret-Ref is a global percentage-point gap between retrieval hit and referral hit. The remaining columns are conditional rates over retrieval misses: each reports the fraction of missed cases in which the agent never issued a tool call matching the corresponding hidden field. Since a single missed case can involve multiple incorrect fields, these diagnostic error rates are not expected to sum to 100\%.

Location is not the dominant retrieval bottleneck for stronger agents. GPT-4.1 and GPT-4.1-mini almost always preserve the correct location when they miss, but schedule, eligibility, intake, and document constraints account for a larger share of retrieval failures. These fields are also the ones that determine whether a referral is practically usable: a pantry in the right county is still a poor match if it is closed at the requested time, requires an unavailable document, or uses an intake method the user cannot satisfy. GPT-OSS-120B and Llama-3.3-70B show a different pattern. Their retrieval misses often involve several fields at once, indicating unstable grounding of the full structured query rather than one consistently difficult slot.

The retrieval-to-referral gap captures a different class of failures. Qwen3-235B retrieves acceptable resources in 94.33\% of cases but recommends them in only 68.67\%, so its main weakness is not search but converting retrieved results into the final referral. DeepSeek-V4-Flash has a smaller but still visible retrieval-to-referral gap despite perfect retrieval. This distinction matters operationally: a food-resource agent can find a valid candidate and still fail the user if the final answer drops the identifier, selects a different resource, or omits the access constraints that made the retrieved resource acceptable.

\begin{table}[t]
\centering
\small
\begin{tabular}{lll}
\toprule
\textbf{Trait} & \textbf{Ref. hit} & \textbf{Retr. hit} \\
\midrule
Normal & 92.0 & 92.0 \\
Unreasonable & \traitcell{90.0}{red}{-2.0} & \traitcell{92.0}{gray}{+0.0} \\
Rambling & \traitcell{86.0}{red}{-6.0} & \traitcell{90.0}{red}{-2.0} \\
Impatience & \traitcell{90.0}{red}{-2.0} & \traitcell{90.0}{red}{-2.0} \\
Incomplete & \traitcell{88.0}{red}{-4.0} & \traitcell{90.0}{red}{-2.0} \\
Inconsistent & \traitcell{82.0}{red}{-10.0} & \traitcell{82.0}{red}{-10.0} \\
\bottomrule
\end{tabular}
\caption{GPT-OSS-120B referral and retrieval hit by interaction trait. Colored deltas show percentage-point changes relative to normal users.}
\label{tab:trait-results}
\end{table}

\begin{table}[t]
\centering
\small
\begin{tabular}{lll}
\toprule
\textbf{Trait} & \textbf{Turns} & \textbf{Tools} \\
\midrule
Normal & 3.18 & 1.28 \\
Unreasonable & \traitcell{3.16}{red}{-0.02} & \traitcell{1.32}{green!50!black}{+0.04} \\
Rambling & \traitcell{3.18}{gray}{+0.00} & \traitcell{1.32}{green!50!black}{+0.04} \\
Impatience & \traitcell{3.30}{green!50!black}{+0.12} & \traitcell{1.24}{red}{-0.04} \\
Incomplete & \traitcell{3.48}{green!50!black}{+0.30} & \traitcell{1.20}{red}{-0.08} \\
Inconsistent & \traitcell{3.16}{red}{-0.02} & \traitcell{1.22}{red}{-0.06} \\
\bottomrule
\end{tabular}
\caption{GPT-OSS-120B interaction efficiency by user trait. Colored deltas show changes relative to normal users.}
\label{tab:trait-efficiency}
\vspace{-10pt}
\end{table}

\subsection{RQ3: Effects of Non-Ideal User Interactions}

We next examine how non-ideal interaction traits affect referral performance and interaction efficiency using GPT-OSS-120B as a representative open-weight agent. Table~\ref{tab:trait-results} reports referral and retrieval hit by interaction trait for this model. The results do not simply rank users by cooperativeness. Normal users are easiest, with 92.0\% referral hit, but the largest drop comes from inconsistent information rather than impatience or unreasonable demands. Inconsistent answers reduce both referral and retrieval hit by 10 percentage points, indicating that same-reply contradictions often enter the search state unresolved. Rambling has a smaller effect on retrieval than on final referral, which suggests that the model can often form a usable query but becomes less reliable when converting a noisier dialogue into a final recommendation.

Table~\ref{tab:trait-efficiency} reports the corresponding interaction efficiency metrics. Incomplete answers increase the conversation length the most, adding 0.30 turns on average, because the first answer to one target fact is intentionally partial. This cost is visible but not catastrophic. Inconsistent users, by contrast, do not increase turns; the model often proceeds with fewer tool calls and lower accuracy. The failure is therefore not that contradictions make the dialogue longer, but that the agent searches before resolving them. Unreasonable demands leave retrieval unchanged but create a small retrieval-to-referral gap, consistent with the agent finding an acceptable food resource while letting the impossible request affect the final response.

These trait results separate three costs of non-ideal user interactions that a single aggregate score would obscure: clarification under contradiction, extra elicitation under partial answers, and final-answer fidelity after noisy but factual dialogue. The same hidden food need remains recoverable across traits, so the benchmark does not rely on impossible tasks; instead, it tests whether the agent preserves and repairs the information needed for a valid referral.

\section{Conclusion}

We introduced \benchmark{}, a food-resource referral framework and benchmark grounded in 686 filtered Indiana food resources. The central challenge is not only whether an agent can call a search tool, but whether it can decide which user statements should become hard filters, which missing facts require follow-up, and which noisy or impossible statements should be ignored when making a referral. Our experiments show that strong LLM agents can perform well in this setting, but their errors are concentrated in constraint grounding and in the handoff from retrieved resources to final recommendations. These failures are exactly the points where a conversational food-referral agent would need the most scrutiny before use outside a controlled evaluation.

\section*{Limitations}

\benchmark{} is a research framework and evaluation benchmark, not a live referral service. The resource index is built from a fixed Indiana 211 source snapshot and then filtered for structured evaluation, so it may omit useful providers and may not reflect current hours, capacity, eligibility rules, or intake procedures. The benchmark currently focuses on Indiana food resources and four food service types, which supports careful tool design but limits geographic and domain generality. The case generator intentionally keeps acceptable-resource sets small enough for automatic scoring, which simplifies real referral practice where several imperfect options may be reasonable. The simulated user model is also an approximation: its traits stress agent behavior, but they cannot capture the full range of lived experience, crisis communication, language access needs, disability access needs, or trust dynamics involved in real food assistance navigation.

\section*{Ethical Considerations}

Food assistance referral is a high-stakes domain involving people who may face poverty, food insecurity, health needs, disability, caregiving burdens, or other vulnerabilities. \benchmark{} is intended only for evaluation and should not be treated as a live referral system. Its resources are filtered for experimental use and may not reflect current availability, eligibility rules, or intake procedures. Any deployment of referral agents should include human oversight, provider validation, escalation paths, crisis-handling mechanisms, language access support, and ongoing monitoring for disparate impacts. \benchmark{} uses simulated users rather than real help-seeking individuals, reducing privacy risks and avoiding exposure of vulnerable users to experimental systems. However, simulation can also underrepresent the emotional, cultural, linguistic, and practical realities of seeking food assistance. We therefore frame \benchmark{} as a tool for identifying agent limitations rather than certifying deployment readiness, and encourage participatory evaluation with community organizations before real-world use.

\bibliography{latex/custom}

@misc{usda2026foodsecurity,
  author       = {{U.S. Department of Agriculture}},
  title        = {Food Security in the U.S.: Key Statistics \& Graphics},
  year         = {2026},
  howpublished = {Web page},
  url          = {https://www.ers.usda.gov/topics/food-nutrition-assistance/food-security-in-the-us/key-statistics-graphics},
  note         = {Accessed: 2026-05-13}
}

@misc{usda2025snap,
  author       = {{U.S. Department of Agriculture}},
  title        = {Supplemental Nutrition Assistance Program (SNAP): Key Statistics and Research},
  year         = {2025},
  month        = jul,
  day          = {24},
  howpublished = {Web page},
  url          = {https://www.ers.usda.gov/topics/food-nutrition-assistance/supplemental-nutrition-assistance-program-snap/key-statistics-and-research},
  note         = {Accessed: 2026-05-13}
}

@article{kreuter2012use,
  title={Use of cancer control referrals by 2-1-1 callers: a randomized trial},
  author={Kreuter, Matthew W and Eddens, Katherine S and Alcaraz, Kassandra I and Rath, Suchitra and Lai, Choi and Caito, Nikki and Greer, Regina and Bridges, Nikisha and Purnell, Jason Q and Wells, Anjanette and others},
  journal={American journal of preventive medicine},
  volume={43},
  number={6},
  pages={S425--S434},
  year={2012},
  publisher={Elsevier}
}

@article{thompson2016promoting,
  title={Promoting health by addressing basic needs: effect of problem resolution on contacting health referrals},
  author={Thompson, Tess and Kreuter, Matthew W and Boyum, Sonia},
  journal={Health Education \& Behavior},
  volume={43},
  number={2},
  pages={201--207},
  year={2016},
  publisher={SAGE Publications Sage CA: Los Angeles, CA}
}

@article{anuyah2023exploring,
  title={Exploring the barriers and potential opportunities of technology integration in Community-Based social service organizations},
  author={Anuyah, Oghenemaro and Conrado, Ann-Marie and Carlson, Clinton and Gilbride, Hope and Metoyer, Ronald},
  journal={ACM Journal on Computing and Sustainable Societies},
  volume={1},
  number={1},
  pages={1--28},
  year={2023},
  publisher={ACM New York, NY}
}

@misc{mcclain_bishop_2026_digital_divides_us,
  author       = {McClain, Colleen and Bishop, William},
  title        = {What We Know About Internet Use, Smartphone Ownership and Digital Divides in the U.S.},
  year         = {2026},
  month        = jan,
  day          = {8},
  institution  = {Pew Research Center},
  type         = {Short Read},
  url          = {https://www.pewresearch.org/short-reads/2026/01/08/internet-use-smartphone-ownership-digital-divides-in-u-s/},
  urldate      = {2026-05-13}
}

@article{bickmore2016improving,
  title={Improving access to online health information with conversational agents: a randomized controlled experiment},
  author={Bickmore, Timothy W and Utami, Dina and Matsuyama, Robin and Paasche-Orlow, Michael K},
  journal={Journal of medical Internet research},
  volume={18},
  number={1},
  pages={e1},
  year={2016},
  publisher={JMIR Publications Inc. Toronto, Canada}
}

@inproceedings{kocielnik2020harborbot,
  title={HarborBot: a chatbot for social needs screening},
  author={Kocielnik, Rafal and Agapie, Elena and Argyle, Alexander and Hsieh, Dennis T and Yadav, Kabir and Taira, Breena and Hsieh, Gary},
  booktitle={AMIA Annual Symposium Proceedings},
  volume={2019},
  pages={552},
  year={2020}
}

@inproceedings{kocielnik2021can,
  title={Can I talk to you about your social needs? Understanding preference for conversational user interface in health},
  author={Kocielnik, Rafal and Langevin, Raina and George, James S and Akenaga, Shota and Wang, Amelia and Jones, Darwin P and Argyle, Alexander and Fockele, Callan and Anderson, Layla and Hsieh, Dennis T and others},
  booktitle={Proceedings of the 3rd Conference on Conversational User Interfaces},
  pages={1--10},
  year={2021}
}

@misc{wa2112023expansion,
  author       = {{Washington 211}},
  title        = {Washington 211 Year End Update: 211 Seeks to Expand Services},
  year         = {2023},
  month        = jan,
  day          = {4},
  howpublished = {Web page},
  url          = {https://wa211.org/washington-211-year-end-update-211-seeks-to-expand-services/},
  note         = {Accessed: 2026-05-13}
}

@article{schick2023toolformer,
  title={Toolformer: Language models can teach themselves to use tools},
  author={Schick, Timo and Dwivedi-Yu, Jane and Dess{\`\i}, Roberto and Raileanu, Roberta and Lomeli, Maria and Hambro, Eric and Zettlemoyer, Luke and Cancedda, Nicola and Scialom, Thomas},
  journal={Advances in neural information processing systems},
  volume={36},
  pages={68539--68551},
  year={2023}
}

@inproceedings{li2023api,
  title={Api-bank: A comprehensive benchmark for tool-augmented llms},
  author={Li, Minghao and Zhao, Yingxiu and Yu, Bowen and Song, Feifan and Li, Hangyu and Yu, Haiyang and Li, Zhoujun and Huang, Fei and Li, Yongbin},
  booktitle={Proceedings of the 2023 conference on empirical methods in natural language processing},
  pages={3102--3116},
  year={2023}
}

@inproceedings{qin2024toolllm,
  title={Toolllm: Facilitating large language models to master 16000+ real-world apis},
  author={Qin, Yujia and Liang, Shihao and Ye, Yining and Zhu, Kunlun and Yan, Lan and Lu, Yaxi and Lin, Yankai and Cong, Xin and Tang, Xiangru and Qian, Bill and others},
  booktitle={International Conference on Learning Representations},
  volume={2024},
  pages={9695--9717},
  year={2024}
}

@article{patil2024gorilla,
  title={Gorilla: Large language model connected with massive apis},
  author={Patil, Shishir G and Zhang, Tianjun and Wang, Xin and Gonzalez, Joseph E},
  journal={Advances in Neural Information Processing Systems},
  volume={37},
  pages={126544--126565},
  year={2024}
}

@inproceedings{alkhouli2025confetti,
  title={CONFETTI: Conversational function-calling evaluation through turn-level interactions},
  author={Alkhouli, Tamer and Margatina, Katerina and Gung, James and Shu, Raphael and Zaghi, Claudia and Sunkara, Monica and Zhang, Yi},
  booktitle={Proceedings of the 63rd Annual Meeting of the Association for Computational Linguistics (Volume 1: Long Papers)},
  pages={7993--8006},
  year={2025}
}

@article{shim2025non,
  title={Non-Collaborative User Simulators for Tool Agents},
  author={Shim, Jeonghoon and Song, Woojung and Jin, Cheyon and KooK, Seungwon and Jo, Yohan},
  journal={arXiv preprint arXiv:2509.23124},
  year={2025}
}

@article{yao2024tau,
  title={\texttt{\textbackslash tau}-bench: A Benchmark for Tool-Agent-User Interaction in Real-World Domains},
  author={Yao, Shunyu and Shinn, Noah and Razavi, Pedram and Narasimhan, Karthik},
  journal={arXiv preprint arXiv:2406.12045},
  year={2024}
}

@inproceedings{lu2025toolsandbox,
  title={Toolsandbox: A stateful, conversational, interactive evaluation benchmark for llm tool use capabilities},
  author={Lu, Jiarui and Holleis, Thomas and Zhang, Yizhe and Aumayer, Bernhard and Nan, Feng and Bai, Haoping and Ma, Shuang and Ma, Shen and Li, Mengyu and Yin, Guoli and others},
  booktitle={Findings of the Association for Computational Linguistics: NAACL 2025},
  pages={1160--1183},
  year={2025}
}

@article{sezgin2024chatbot,
  title={Chatbot for social need screening and resource sharing with vulnerable families: iterative design and evaluation study},
  author={Sezgin, Emre and Kocaballi, A Baki and Dolce, Millie and Skeens, Micah and Militello, Lisa and Huang, Yungui and Stevens, Jack and Kemper, Alex R},
  journal={JMIR Human Factors},
  volume={11},
  number={1},
  pages={e57114},
  year={2024},
  publisher={JMIR Publications Inc., Toronto, Canada}
}

@misc{openai2025gpt41,
  author       = {{OpenAI}},
  title        = {Introducing {GPT-4.1} in the {API}},
  year         = {2025},
  month        = apr,
  day          = {14},
  howpublished = {Blog post},
  url          = {https://openai.com/index/gpt-4-1/},
  note         = {Accessed: 2026-05-14}
}

@misc{openai2025gptoss,
  author       = {{OpenAI}},
  title        = {{gpt-oss-120b} \& {gpt-oss-20b} Model Card},
  year         = {2025},
  month        = aug,
  day          = {5},
  howpublished = {Model card},
  url          = {https://openai.com/index/gpt-oss-model-card/},
  note         = {Accessed: 2026-05-14}
}

@misc{yang2025qwen3,
  title         = {{Qwen3} Technical Report},
  author        = {Yang, An and Li, Anfeng and Yang, Baosong and Zhang, Beichen and Hui, Binyuan and Zheng, Bo and Yu, Bowen and Gao, Chang and Huang, Chengen and Lv, Chenxu and others},
  year          = {2025},
  eprint        = {2505.09388},
  archivePrefix = {arXiv},
  primaryClass  = {cs.CL},
  url           = {https://arxiv.org/abs/2505.09388}
}

@misc{meta2024llama33,
  author       = {{Meta AI}},
  title        = {{Llama 3.3}: Model Cards and Prompt Formats},
  year         = {2024},
  howpublished = {Model card},
  url          = {https://www.llama.com/docs/model-cards-and-prompt-formats/llama3_3/},
  note         = {Accessed: 2026-05-14}
}

@misc{feedingamerica_network,
  author       = {{Feeding America}},
  title        = {Our Work},
  howpublished = {\url{https://www.feedingamerica.org/our-work}},
  note         = {Accessed: 2026-05-31}
}

@article{ye2025llms4all,
  title={Llms4all: A review of large language models across academic disciplines},
  author={Ye, Yanfang and Zhang, Zheyuan and Ma, Tianyi and Wang, Zehong and Li, Yiyang and Hou, Shifu and Sun, Weixiang and Shi, Kaiwen and Ma, Yijun and Song, Wei and others},
  journal={arXiv preprint arXiv:2509.19580},
  year={2025}
}

@inproceedings{chen2025obvious,
  title={The obvious invisible threat: Llm-powered gui agents’ vulnerability to fine-print injections},
  author={Chen, Chaoran and Zhang, Zhiping and Guo, Bingcan and Ma, Shang and Khalilov, Ibrahim and Gebreegziabher, Simret and Ye, Yanfang and Xiao, Ziang and Yao, Yaxing and Li, Tianshi and others},
  booktitle={Soups},
  year={2025},
  organization={The Twenty-First Symposium on Usable Privacy and Security (SOUPS)}
}

@inproceedings{chen2025clear,
  title={Clear: Towards contextual llm-empowered privacy policy analysis and risk generation for large language model applications},
  author={Chen, Chaoran and Zhou, Daodao and Ye, Yanfang and Li, Toby Jia-jun and Yao, Yaxing},
  booktitle={Proceedings of the 30th International Conference on Intelligent User Interfaces},
  pages={277--297},
  year={2025}
}

@inproceedings{zhao2023self,
  title={Self-supervised graph structure refinement for graph neural networks},
  author={Zhao, Jianan and Wen, Qianlong and Ju, Mingxuan and Zhang, Chuxu and Ye, Yanfang},
  booktitle={Proceedings of the sixteenth ACM international conference on web search and data mining},
  pages={159--167},
  year={2023}
}

@inproceedings{chen2018droideye,
  title={Droideye: Fortifying security of learning-based classifier against adversarial android malware attacks},
  author={Chen, Lingwei and Hou, Shifu and Ye, Yanfang and Xu, Shouhuai},
  booktitle={2018 IEEE/ACM International Conference on Advances in Social Networks Analysis and Mining (ASONAM)},
  pages={782--789},
  year={2018},
  organization={IEEE}
}

@inproceedings{ye2009intelligent,
  title={Intelligent file scoring system for malware detection from the gray list},
  author={Ye, Yanfang and Li, Tao and Jiang, Qingshan and Han, Zhixue and Wan, Li},
  booktitle={Proceedings of the 15th ACM SIGKDD international conference on Knowledge discovery and data mining},
  pages={1385--1394},
  year={2009}
}

@inproceedings{ju2022grape,
  title={Grape: Knowledge graph enhanced passage reader for open-domain question answering},
  author={Ju, Mingxuan and Yu, Wenhao and Zhao, Tong and Zhang, Chuxu and Ye, Yanfang},
  booktitle={Findings of the Association for Computational Linguistics: EMNLP 2022},
  pages={169--181},
  year={2022}
}

@article{qian2022co,
  title={Co-modality graph contrastive learning for imbalanced node classification},
  author={Qian, Yiyue and Zhang, Chunhui and Zhang, Yiming and Wen, Qianlong and Ye, Yanfang and Zhang, Chuxu},
  journal={Advances in Neural Information Processing Systems},
  volume={35},
  pages={15862--15874},
  year={2022}
}

@inproceedings{zhao2021multi,
  title={Multi-view self-supervised heterogeneous graph embedding},
  author={Zhao, Jianan and Wen, Qianlong and Sun, Shiyu and Ye, Yanfang and Zhang, Chuxu},
  booktitle={Joint European conference on machine learning and knowledge discovery in databases},
  pages={319--334},
  year={2021},
  organization={Springer}
}

@article{ju2023graphpatcher,
  title={Graphpatcher: mitigating degree bias for graph neural networks via test-time augmentation},
  author={Ju, Mingxuan and Zhao, Tong and Yu, Wenhao and Shah, Neil and Ye, Yanfang},
  journal={Advances in Neural Information Processing Systems},
  volume={36},
  pages={55785--55801},
  year={2023}
}

\appendix

\section{Additional Benchmark Details}
\label{sec:appendix-details}

\subsection{Resource Processing Details}
\label{sec:appendix-resource-processing}

Table~\ref{tab:appendix-filtering} shows the resource filtering pipeline used to construct the filtered food index. The filtering steps keep records whose food type, location, schedule, eligibility, intake, document, and referral metadata can be represented by the tool schema.

\begin{table}[h]
\centering
\small
\begin{tabular}{lrr}
\toprule
\textbf{Step} & \textbf{Kept} & \textbf{Removed} \\
\midrule
Processed food resources & 2,422 & -- \\
Structured schedule & 1,422 & 1,000 \\
Known food service type & 1,348 & 74 \\
Benchmark food type & 1,026 & 322 \\
Known eligibility tags & 942 & 84 \\
Parseable intake methods & 868 & 74 \\
Parseable documents & 686 & 182 \\
Benchmark metadata & 686 & 0 \\
\bottomrule
\end{tabular}
\caption{Filtering pipeline for the filtered food resource index.}
\label{tab:appendix-filtering}
\end{table}

The final tagged resource index keeps \texttt{resource\_id}, \texttt{agency\_name}, \texttt{site\_name}, \texttt{service\_name}, \texttt{service\_track}, \texttt{food\_service\_types}, \texttt{eligibility\_tags}, \texttt{service\_area}, \texttt{city}, \texttt{state}, \texttt{zipcode}, \texttt{address\_1}, \texttt{phone}, \texttt{website}, \texttt{schedule\_status}, \texttt{schedule\_windows}, \texttt{intake\_methods}, \texttt{document\_requirements}, and \texttt{eligibility\_text}. Table~\ref{tab:appendix-distribution} reports the main tagged distributions in the final index.

\begin{table*}[t]
\centering
\small
\begin{tabular}{p{0.22\textwidth}rp{0.25\textwidth}rp{0.22\textwidth}r}
\toprule
\textbf{Food type} & \textbf{Count} & \textbf{Eligibility tag} & \textbf{Count} & \textbf{Intake method} & \textbf{Count} \\
\midrule
Food pantry/groceries & 485 & Resident/service area & 518 & Walk in & 480 \\
Prepared meal & 160 & Open eligibility & 135 & Call & 130 \\
Home-delivered meals & 28 & Senior & 60 & Drive-through & 103 \\
Baby food/formula & 24 & Low income & 48 & Appointment & 83 \\
 & & Homebound/disabled & 39 & & \\
 & & Child/student & 8 & & \\
 & & Pregnant/WIC & 4 & & \\
 & & Veteran & 1 & & \\
\bottomrule
\end{tabular}
\caption{Tagged distribution in the filtered food resource index. Counts can exceed 686 because resources may have multiple tags.}
\label{tab:appendix-distribution}
\end{table*}

\subsection{Tool Schema and Prompts}
\label{sec:appendix-tool-agent}

The benchmark exposes a single function tool, \texttt{search\_food\_resources}. Table~\ref{tab:appendix-tool-schema} gives the schema shown to the agent. Non-empty fields are interpreted as hard filters, multiple values inside one field are interpreted disjunctively, and ZIP-code filters take precedence over cities and counties.

\begin{table*}[t]
\centering
\small
\begin{tabular}{p{0.22\textwidth}p{0.24\textwidth}p{0.45\textwidth}}
\toprule
\textbf{Argument} & \textbf{Type / values} & \textbf{Description} \\
\midrule
\texttt{counties} & list of strings & County service-area filter, using uppercase county names. \\
\texttt{cities} & list of strings & City filter. \\
\texttt{zipcodes} & list of strings & Five-digit ZIP-code filter. \\
\texttt{service\_tracks} & enum list & Broad food track. \\
\texttt{food\_service\_types} & enum list & Food type such as pantry/groceries, prepared meal, home-delivered meals, or baby food/formula. \\
\texttt{eligibility\_tags} & enum list & User facts such as senior, child/student, pregnant/WIC, homebound/disabled, veteran, low income, resident/service area, or open eligibility. \\
\texttt{intake\_methods} & enum list & Firm access or intake method such as walk-in, call, drive-through, or appointment. \\
\texttt{available\_documents} & enum list & Documents the user can provide; a resource matches when its document requirements are a subset of this list. \\
\texttt{schedule} & object & Day plus daypart, 24-hour, or appointment-only schedule filter. \\
\texttt{include\_appointment\_only} & boolean & Controls whether appointment-only resources are included when no explicit schedule filter is used. \\
\texttt{limit} & integer, 1--20 & Maximum number of returned resources. \\
\bottomrule
\end{tabular}
\caption{Arguments of the \texttt{search\_food\_resources} tool.}
\label{tab:appendix-tool-schema}
\end{table*}

The tool returns a JSON object with a \texttt{resources} list. Each returned resource includes the resource identifier, resource name, agency name, food service type tags, eligibility tags, city, ZIP code, contact fields, schedule status, intake methods, document requirements, and eligibility text.

The core agent instruction prompt is:

\begin{PromptBlock}
You are a food-focused Indiana 211 resource retrieval agent. Your job is to
understand the user's food situation, call the search tool, and make one final
resource recommendation.

Before searching, clarify the food help type, one location constraint, schedule,
access/intake constraints, document availability, and any relevant program
qualification facts. Ask one question at a time.

Do not ask the user for "eligibility tags" or whether they are "eligible" in the
abstract. Ask in ordinary language for facts that food programs commonly use,
such as whether the help is for a senior or older adult, a child or student, someone
who is pregnant or using WIC, someone homebound or disabled, a veteran, or someone
with limited income. Use the user's
answers to choose the appropriate `eligibility_tags` in the search call.

Use access/intake and document fields as search filters only when the user gives
a positive firm constraint. If the user has no special access need or no
specific document constraint, omit those fields from the search call.

Call the search tool at most three times total. The final answer must be strict
JSON only:

{
  "recommendations": [
    {
      "resource_name": "Resource name from the tool",
      "resource_id": "full in211-... id from the tool",
      "intake_methods": [],
      "document_requirements": []
    }
  ]
}

Recommend up to three resources. Copy resource fields exactly from tool results.
If no suitable resource is found, output {"recommendations": []}.
\end{PromptBlock}

The user-card generation prompt receives a deterministic case specification and asks an LLM to create a natural hidden card without adding new constraints. The most important rules are:

\begin{PromptBlock}
You create hidden simulated-user cards for a food-focused Indiana 211 benchmark.

Convert the deterministic user spec into a realistic hidden user card. Return
strict JSON only.

Rules:
- Create one plain-language `need_summary` for each food need.
- In `opening`, do not mention time, urgency, eligibility, documents, intake
  method, access method, transportation, or resource preferences.
- `trait_openings.unreasonable_demand` must exaggerate the same underlying food
  need into an unrealistic demand without adding unrelated needs.
- `persona` is only a voice cue. It must not add hidden constraints.
- Treat location, schedule, food type, eligibility, intake, and document facts
  in the deterministic spec as fixed truth.
- Do not mention resource IDs, agency names, site names, or specific resource names.
\end{PromptBlock}

\subsection{User Simulation Prompts and Trait Templates}
\label{sec:appendix-user-simulation}

The simulated user is prompted to act as a real person seeking help from Indiana 211 while following a hidden user card. The hidden card contains the user's persona, location, fixed location requirement, food need, schedule, eligibility facts, and any firm intake or document constraints. The user is instructed to answer naturally from the hidden facts, provide the fixed location requirement when asked about location, and avoid creating unrelated new needs or constraints.

The full simulated-user instruction prompt is:

\begin{PromptBlock}
You are simulating a real person seeking help from Indiana 211.

You must follow the hidden user context. The hidden context contains the user's
persona, location background, needs, and firm constraints.

The hidden `location_requirement` is fixed truth. Each item in hidden `needs`
has a plain-language need summary, food help type, eligibility facts, and a
firm schedule requirement. In this benchmark, composite cases are still one
food problem; they simply include extra firm access/intake or document facts.

Only `location_requirement` and the fields inside hidden `needs` define hard
constraints. The hidden `persona` is only a voice cue for realism; do not turn
it into new location, access, document, schedule, eligibility, or food-type
requirements.

Trait behavior changes communication style only; it does not change the hidden
truth. The only exception is an explicit `Trait state` instruction for
incomplete_answer or inconsistency, and that exception applies only to the
selected target fact and only for the current reply.

When asked about program qualification facts, answer only with qualification
facts listed in the hidden needs. Do not infer low income, disability,
pregnancy/WIC, senior status, student status, veteran status, or other
qualification facts from general background unless that eligibility tag is
present in the hidden needs.

When the agent asks about location or where resources should be searched, state
your own location information naturally and your `location_requirement` clearly.
However, if your trait defines a reply style, prioritize that style when responding.

If the agent asks about a constraint not present in the hidden facts, answer
naturally that you have no special requirement.

If the agent asks about access/intake or documents and those facts are present
in the hidden need, answer naturally but treat them as firm constraints. If a
fact is not present in the hidden need, say naturally that there is no special
requirement and do not add opinions or side comments about that category.

Speak casually and naturally, like a real person asking for help. However,
prioritize adjusting your wording to align with the trait definition while
keeping the underlying need and identity consistent.

Do not create unrelated new service needs, errands, or tasks beyond the hidden
profile.
\end{PromptBlock}

The trait instructions used by the simulated user are:
\begin{itemize}
    \item \textbf{Normal:} be cooperative and answer questions clearly and directly.
    \item \textbf{Unreasonable demand:} demand an unrealistic food-related outcome, such as grocery cash or instant delivery, while keeping the recoverable food need recognizable.
    \item \textbf{Rambling responses:} answer the factual question truthfully but surround the answer with side comments that do not add constraints.
    \item \textbf{Impatience:} be rude, rushed, and frustrated while still providing enough factual information for the conversation to continue.
    \item \textbf{Incomplete answers:} give an unclear or partial answer only for the selected target fact the first time that fact is directly requested.
    \item \textbf{Inconsistent information:} give a self-contradictory answer only for the selected target fact and only within one reply, so the agent has evidence that clarification is needed.
\end{itemize}

Each simulated-user card contains the fields shown in Table~\ref{tab:appendix-user-card}. The ground-truth resource identifiers are included in the card file for scoring but are not revealed to the agent.

\begin{table}[h]
\centering
\small
\begin{tabular}{p{0.39\columnwidth}p{0.50\columnwidth}}
\toprule
\textbf{Field} & \textbf{Purpose} \\
\midrule
\texttt{case\_id}, \texttt{user\_id} & Stable identifiers. \\
\texttt{case\_type} & Single-need or composite-need case. \\
\texttt{difficulty}, \texttt{difficulty\_factors} & Deterministic case difficulty metadata. \\
\texttt{traits} & Interaction trait assigned to the user. \\
\texttt{opening} & First user message. \\
\texttt{persona} & Short hidden voice cue for realistic phrasing. \\
\texttt{location} & User-facing location information. \\
\texttt{location\_requirement} & Hidden structured location constraint. \\
\texttt{needs} & Hidden food need, food type, eligibility, schedule, and optional intake or document constraints. \\
\texttt{acceptable\_resource\_ids} & Acceptable resource identifiers used only for scoring. \\
\texttt{acceptable\_resources} & Acceptable resource metadata used only for scoring and analysis. \\
\bottomrule
\end{tabular}
\caption{Fields used to prompt the simulated user.}
\label{tab:appendix-user-card}
\end{table}

An example full simulated-user card is:

\begin{PromptBlock}
{
  "case_id": "food-spec-008",
  "user_id": "food-spec-008__normal",
  "case_type": "composite",
  "difficulty": "hard",
  "difficulty_factors": [
    "single_acceptable_resource",
    "specific_daypart",
    "intake_constraint",
    "composite_access_or_document_constraint"
  ],
  "traits": ["normal"],
  "opening": "I'm looking for places where I can get a hot meal in Marion County, Indiana.",
  "persona": "I'm a single adult living in Indianapolis.",
  "location": {"county": "MARION", "city": "Indianapolis", "state": "IN", "zipcode": "46241"},
  "location_requirement": {"counties": ["MARION"]},
  "needs": [
    {
      "need_id": "need_1",
      "need_summary": "I need a place nearby where I can get a freshly prepared meal.",
      "service_tracks": ["direct_food"],
      "food_service_types": ["prepared_meal"],
      "eligibility_tags": ["resident_area"],
      "schedule": {"day": "wed", "time": "evening"},
      "intake_methods": ["drive_through"]
    }
  ],
  "acceptable_resource_ids": ["in211-3082-75313-community-meal"],
  "acceptable_resources": [
    {
      "resource_id": "in211-3082-75313-community-meal",
      "service_name": "Community Meal",
      "agency_name": "Lynhurst Baptist Church",
      "food_service_types": ["prepared_meal"],
      "eligibility_tags": ["resident_area"],
      "intake_methods": ["drive_through"],
      "document_requirements": []
    }
  ]
}
\end{PromptBlock}

\subsection{Benchmark Case Construction}
\label{sec:appendix-case-construction}

Case construction starts from candidate resources in the filtered index. The builder enumerates possible location constraints from county, city, and ZIP-code metadata and schedule constraints from parsed schedule windows. It retains a case only when the hidden constraints retrieve one to three acceptable resources. Single cases use food type, service track, eligibility, location, and schedule constraints. Composite cases add a firm intake or document constraint to the same food problem.

The validation step enforces the following invariants:
\begin{itemize}
    \item every case has one food need and exactly one location requirement;
    \item every case has a food service type, service track, eligibility facts, and one schedule requirement;
    \item acceptable resources are produced by the same tool semantics used at evaluation time;
    \item each case has between one and three acceptable resources;
    \item composite cases add at least one firm intake or document constraint;
    \item the simulated-user opening does not reveal hidden schedule, eligibility, intake, or document facts.
\end{itemize}

The resulting benchmark has 50 seed cases and 300 evaluation samples. It has 25 single seed cases and 25 composite seed cases. Location requirements use ZIP codes in 18 seed cases, cities in 16 seed cases, and counties in 16 seed cases. All seed cases use a day plus daypart schedule requirement.

\section{Complete Qualitative Examples}
\label{sec:appendix-examples}

Example~\ref{sec:appendix-examples} shows a successful composite episode. The case is composite because it includes a firm drive-through intake constraint in addition to the core food need.

\begin{PromptBlock}
Case: food-spec-008__normal
Ground truth:
- food_service_types=["prepared_meal"]
- eligibility_tags=["resident_area"]
- counties=["MARION"]
- schedule={day:"wed", time:"evening"}
- intake_methods=["drive_through"]
- acceptable_resource_id="in211-3082-75313-community-meal"

User: I'm looking for places where I can get a hot meal in Marion County, Indiana.

Agent: To help find the best options for you, could you share a bit more about when you would like to get a hot meal? For example, do you need a meal today, on weekdays, weekends, or at a specific time of day?

User: I need a freshly prepared meal on Wednesday evening.

Agent: Thank you! Is there anything else you need to consider, such as needing wheelchair access, preferring walk-in options only, or needing to avoid locations that require you to show ID or other documents? Or are you open to any available options?

User: I specifically need a place that offers drive-through service. Other than that, I'm open to any options.

Tool call:
search_food_resources({
  "counties": ["MARION"],
  "cities": [],
  "zipcodes": [],
  "service_tracks": ["direct_food"],
  "food_service_types": ["prepared_meal"],
  "eligibility_tags": [],
  "intake_methods": ["drive_through"],
  "available_documents": [],
  "schedule": {"day": "wed", "time": "evening", "requires_24_hours": false, "appointment_only": false},
  "include_appointment_only": false,
  "limit": 3
})

Tool result:
{
  "resources": [
    {
      "resource_id": "in211-3082-75313-community-meal",
      "resource_name": "Community Meal",
      "agency_name": "Lynhurst Baptist Church",
      "food_service_types": ["prepared_meal"],
      "eligibility_tags": ["resident_area"],
      "city": "Indianapolis",
      "zipcode": "46241",
      "intake_methods": ["drive_through"],
      "document_requirements": []
    }
  ]
}

Agent final answer:
{
  "recommendations": [
    {
      "resource_name": "Community Meal",
      "resource_id": "in211-3082-75313-community-meal",
      "intake_methods": ["drive_through"],
      "document_requirements": []
    }
  ]
}
\end{PromptBlock}

\end{document}